\documentclass[10pt,twocolumn,letterpaper]{article}

\usepackage[pagenumbers]{wacv} 

\usepackage{graphicx}
\usepackage{amsmath}
\usepackage{amssymb}
\usepackage{booktabs}
\usepackage{xcolor}
\usepackage{multirow}
\usepackage[normalem]{ulem}
\usepackage{makecell}

\usepackage[pagebackref,breaklinks,colorlinks]{hyperref}

\usepackage[capitalize]{cleveref}
\crefname{section}{Sec.}{Secs.}
\Crefname{section}{Section}{Sections}
\Crefname{table}{Table}{Tables}
\crefname{table}{Tab.}{Tabs.}

\newif\ifshowcomment
    \showcommentfalse
\ifshowcomment
    \newcommand{\mc}[1]{\textcolor{olive}{[michael: #1]}}
    \newcommand{\my}[1]{\textcolor{teal}{[meng: #1]}}
    
\else
    \newcommand{\mc}[1]{}
    \newcommand{\my}[1]{}
\fi

\def\wacvPaperID{297}
\def\confName{WACV}
\def\confYear{2025}

\begin{document}

\title{A Video is Worth 10,000 Words: Training and Benchmarking with Diverse Captions for Better Long Video Retrieval}

\author{Matthew Gwilliam$^{1}$\thanks{Work performed during internship with SRI International.} \quad Michael Cogswell$^{2}$ \quad Meng Ye$^{2}$ \quad Karan Sikka$^{2}$ \\ \quad Abhinav Shrivastava$^{1}$\quad Ajay Divakaran$^{2}$\\[0.5em]
$^{1}$University of Maryland, College Park \quad\quad $^{2}$SRI International \quad\quad \\
}

\maketitle

\begin{abstract}
   Existing long video retrieval systems are trained and tested in the paragraph-to-video retrieval regime, where every long video is described by a single long paragraph. This neglects the richness and variety of possible valid descriptions of a video, which could range anywhere from moment-by-moment detail to a single phrase summary. To provide a more thorough evaluation of the capabilities of long video retrieval systems, we propose a pipeline that leverages state-of-the-art large language models to carefully generate a diverse set of synthetic captions for long videos. We validate this pipeline's fidelity via rigorous human inspection. We use synthetic captions from this pipeline to perform a benchmark of a representative set of video language models using long video datasets, and show that the models struggle on shorter captions. We show that finetuning on this data can both mitigate these issues (+2.8\% R@1 over SOTA on ActivityNet with diverse captions), and even improve performance on standard paragraph-to-video retrieval (+1.0\% R@1 on ActivityNet). We also use synthetic data from our pipeline as query expansion in the zero-shot setting (+3.4\% R@1 on ActivityNet). We derive insights by analyzing failure cases for retrieval with short captions.
\end{abstract}

\section{Introduction}
\label{sec:intro}

\begin{figure}[h!]
    \centering
    \includegraphics[width=1.0\linewidth]{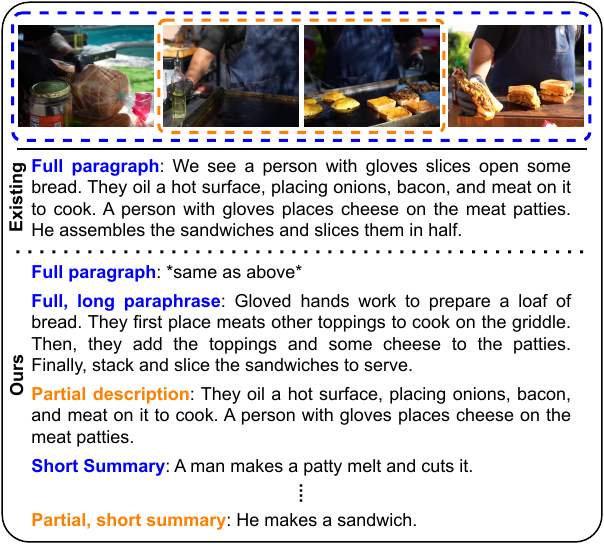}
    \caption{In real-world text-to-video retrieval, users could use diverse queries. 
    Standard long video datasets use only paragraph-style captions (``Existing'', ``Full paragraph''), 
    which does not allow for training 
    or evaluation on a representative set of long video descriptions.
    Practical applications also require the ability to handle complex, 
    short, and partial descriptions of a long video.
    In this work, we introduce an approach to generate, evaluate, 
    and train on such diverse video description data.}
    \vspace{-1.5em}
    \label{fig:teaser}
\end{figure}

If a picture is worth 1,000 words, then a video is worth 10,000.
Consider the variety of possible captions that can describe just one video (Figure~\ref{fig:teaser}).
Although they can vary substantially in semantics and structure, video-language models ought to be able to match all of these captions with the video they describe.

In this paper we show that existing approaches fail to model the variety of captions and show how they can be improved in the context of video retrieval.
At its core, video retrieval requires not just a system that understands video and text, but also how minor differences between videos in a dataset make them unique.
However, video retrieval literature traditionally considers just short clips~\cite{kaufman2017temporal,miech2019howto100m}, which cannot be described by such a variety of captions, and thus obscures the problem.
Increasingly more works have focused on long videos with multiple events,
but it uses only full paragraphs for retrieval~\cite{bain2022frozen,bain2022cliphitchhikers}, neglecting the rich space of valid captions.
While even existing captions can be ambiguous~\cite{Wray_2021_CVPR}, they still
do not include vague, abstract, or partial descriptions a user (\textit{e.g.}, doing video search) might give.
This means current video retrieval datasets do not measure real world performance, 
where captions can be ambiguous, vary in semantics and style, and can describe long complex videos.

To address this we formulate the \textbf{10k Words} benchmark, a novel video retrieval setting which includes diverse descriptions generated for long videos with multiple events.
We identify key axes of variation, including simplification, summarization, and duration, then use them to curate pools of captions with non-trivial differences in structure and semantics.
The benchmark introduces challenging ambiguities, since some captions will not mention all the details that distinguish a video from similar, related videos.
We instantiate this benchmark by augmenting existing datasets~\cite{ghanem2018activitynet,oncescu2021queryd,sun2023longform} with diverse captions, creating ActivityNet10k, QuerYD10k, and LF-VILA10k (borrowing 10k from the idea that ``a video is worth 10,000 words,'' and we work towards that richness of description with our diverse sythetic captions).
These augmentations are only possible given the flexibility and accuracy of recent large language models (LLMs)~\cite{brown2020language}, which we combine with some simple automatic manipulations to synthesize the diverse 10k Words datasets as described in Section~\ref{sec:task_details}.

These proposed datasets can help us detect failures of existing models to capture the space of text descriptions as well as help us to mitigate those failures, and we show both.
For detection, we consider a representative set of state-of-the-art video models and show that they struggle to adequately
solve the 10k Words problem in Section~\ref{sec:benchmark}, struggling especially with short, summary-style captions.
We then demonstrate the effectiveness of a simple mitigation strategy that uses 10k datasets to augment standard datasets during training.
This can provide an inexpensive boost to performance on both the 10k datasets and the original standard datasets, or be used to increase data efficiency.
We also investigate inference time improvement, showing how query expansion can benefit a pre-trained model without finetuning.
An LLM can be used during inference to improve retrieval performance by generating multiple queries for the same video.
We achieve SOTA performance on 10k Words 
while also boosting performance on the standard paragraph-to-video retrieval task.

Finally, in Section~\ref{sec:understanding_failures} we analyze failure cases to understand whether shorter captions are truly ambiguous or not.
We find some cases our not ambiguous, indicating our models have room for improvement on our 10k benchmark.
In summary, we contribute the following:

\begin{itemize}
    \item We instantiate the 10k Words benchmark, a framework for characterizing the broad spectrum of valid descriptions for long videos, by creating ActivityNet10k, QuerYD10k, and LF-VILA10k with a flexible LLM-based pipeline.
    \item We evaluate SOTA models in a zero-shot fashion, and reveal that they struggle on the 10k Words benchmark.
    \item We leverage 10k data for an improvement of +3.4\% R@1 on zero-shot standard ActivityNet retrieval (without finetuning) and +2.8\% R@1 on ActivityNet10k retrieval (with finetuning), which is SOTA 10k Words performance.
    \item We show that despite the ambiguity of shorter captions, SOTA models still fail in non-ambiguous cases.
\end{itemize}

\section{10k Words Benchmark and Datasets}
\label{sec:task_details}

\subsection{10k Definition and Generation}

Given an existing dataset of videos and corresponding descriptions, we create a 10k version of the dataset by enriching the set of descriptions to cover more possible ways to describe the videos.
Existing datasets like those in Table~\ref{tab:datasets}
often take a long video and annotate $E$ events ${e_1, e_2, \ldots, e_E}$ individually.
Each event $e_i$ has a corresponding short video clip $v_i$ and is annotated with a natural language
description of that clip $t_i$, with the set of clips and texts for a given video being denoted
$V$ and $T$.
As such, the original long caption could be a paragraph, long
sentence, or, more typically, the concatenation of video segment captions, which is then treated as a paragraph.
To cover the broadest possible spectrum of natural language queries for a video
we start by defining three augmentation axes along which a video's description can vary:
\textit{duration}, \textit{summarization}, and \textit{simplification}.
Duration refers to how many of the events in a video are described by a given query, while summarization and simplification cover different ways of using language to describe the same video. 
For each axis we implement a function that takes a video with event segmentation and descriptions as input and outputs a new augmented version of the same video with a new set of segments and descriptions.
Prior to LLMs, summarization and simplification would have been difficult to simulate effectively and reliably, and perhaps would have required expensive human annotations.
However, now we are able to effectively prompt LLM to gather such data~\cite{ye2023comprehensive,brown2020language}.
Next we discuss the prompts we design for each augmentation axis, with example synthetic notations in Table~\ref{tab:10k_notation}.

\begin{table*}[t]
    \centering
    \caption{10k Data and Notation. We give our diversity axes, levels, and show an example of the captions for 1 video.}
    \vspace{-0.5em}
    \resizebox{1.\linewidth}{!}{
    \begin{tabular}{@{}lll@{}}
    \toprule
    Axis & Level & Example \\
    \midrule
    Ground Truth & Full (f) & People are sitting in kayaks paddling in the water. They go under a rock and through a tunnel. \\
     & Partial (p) & They go under a rock and through a tunnel. \\
    \midrule
    Summarization & Short (s) & Kayakers paddle, go under rock, through tunnel. \\
     & Medium (m) & People in kayaks paddle, pass under rock, navigate through tunnel in water. \\
     & Long (l) & A group of kayakers paddle through water, passing under a rock and navigating through a tunnel. \\
    \midrule
    Simplification & Elementary (e) & People are in small boats and paddle in the water. They go under a big rock and through a tunnel. \\
     & Intermediate (i) & Individuals are seated in kayaks, using paddles to navigate through the water. They pass beneath \\
     & & a large rock formation and venture through a tunnel. \\
     & University (u) & A group of individuals are situated in kayaks, propelling themselves forward with paddles as they traverse\\
     & & the water. They maneuver beneath a substantial rock structure and proceed through a tunnel. \\
    \bottomrule
    \end{tabular}
    }
    \label{tab:10k_notation}
\end{table*}

\begin{table}[h]
    \begin{minipage}{1.\linewidth}
    \caption{\textbf{Datasets} which we create with our synthetic caption generation pipeline.
    We use only the `train' and `val-1' splits of ActivityNet Captions, and do some additional filtering for extremely long outlier captions from LF-VILA and QuerYD.}
    \label{tab:datasets}
    \vspace{-0.5em}
    \resizebox{1.\textwidth}{!}{
        \setlength{\tabcolsep}{8pt}
        \begin{tabular}{@{}l l c c c@{}} 
            \toprule
            Dataset & Source Dataset & \# Videos & Video Len (s) & Text Len (w) \\
            \midrule
            ActivityNet10k & ActivityNet Captions~\cite{krishna2017densecaptioning}  & 14926 & 117.5 & 49.8 \\
            LF-VILA10k & LF-VILA~\cite{xue2022advancing,sun2023longform} & 7020 & 203.9 & 155.4 \\
            QuerYD10k & QuerYD~\cite{oncescu2021queryd} & 2474 & 264.3 & 203.8 \\
            \bottomrule
        \end{tabular}
    }
    \vspace{-1.5em}
    \end{minipage} 
\end{table}

\textbf{Summarization.}
Descriptions of videos can vary in length.
While at one extreme they describe every detail in the video, at the other they briefly
describe the main idea, leaving out some significant details.
In between the two extremes, relevant details are progressively grouped and redundant elements are pruned.
At one end of this spectrum a video retrieval model must be able to parse details and at the other end it must be able to understand a gestalt.
To augment a video on this axis we prompt an LLM with the ground
truth descriptions $T$ (concatenated) and instruct it to generate summaries.
If the concatenated description has $L$ words, then we ask the LLM to generate three summaries with $\lfloor {L} \cdot \frac{l}{7} \rfloor$ words each
for $l \in \{1, 4, 7\}$.
%
At full length ($l=7$) this should just re-phrase the concatenated caption, but
at smaller lengths the LLM must leave out information.
We observe that GPT-3.5\footnote{\label{gpt3.5}\href{https://platform.openai.com/docs/models/gpt-3-5}{gpt-3.5-turbo-0613}} is able to achieve close to the desired word count most of the time.
This only changes $T$, leaving $E$ and $V$ unchanged.

\textbf{Simplification.}
Descriptions of videos can vary in terms of their conceptual simplicity, where
an idea could be described at the level of a college graduate, or else
simplified for a kindergartener, and a good retrieval model should map all these descriptions to the same video.
We capture this dimension by providing an LLM with the same ground truth
description as for summarization and instruct it to output a simplified
version.
This is done for three levels of reading comprehension described to the LLM as
``elementary'', ``intermediate'', or ``university'' reading level.
This only modifies $T$, leaving $E$ and $V$ unchanged.

\textbf{Duration.}
Descriptions of videos can be partial, intending to cover only a segment of
the video, but the video should still usually be retrieved when these are used as queries (see more about ambiguity in Section~\ref{sec:understanding_failures}).
In our dataset we implement this by choosing a contiguous subset of events
$\tilde{E} = {e_i, \ldots, e_j}$ with start and end index $i$ and $j$.
The corresponding set of video clips $\tilde{V}$ and captions $\tilde{T}$ are
selected to create the augmented video.

\textbf{10k Datasets.}
We combine these axes to construct 10k versions of ActivityNet Captions (ActivityNet), QuerYD, and LF-VILA~(Table~\ref{tab:datasets}). 
%
We construct our 10k Words datasets by taking the per-segment captions available for 
the datasets described in Table~\ref{tab:datasets} and feed them to GPT-3.5 with relevant prompts.
Starting from each video in a base dataset like ActivityNet, we include 11 captions 
for each video:
1 \textbf{f}ull caption (original ground truth paragraph), 3 captions for the levels of
simplification (\textbf{e}lementary, \textbf{i}ntermediate, and \textbf{u}niversity), 3 captions for the 
levels of summarization (\textbf{s}hort, \textbf{m}edium, and \textbf{l}ong), 3 captions
that combine summarization and simplification by generating simplifications for
the \textbf{s}hort summaries,
and 1 caption corresponding to a random subset of the original video segments by duration augmentation.
We show examples and 
introduce relevant shorthand in Table~\ref{tab:10k_notation}.

We refer to the 10k Words version of LF-VILA, a sample from the original LF-VILA~\cite{sun2023longform},
as LF-VILA10k.
We use all of QuerYD as a validation set, since the initial small size of its validation 
set makes it challenging to distill useful insights, and create QuerYD10k. 
We create ActivityNet10k for ActivityNet.
We provide details on LLM prompts and costs in the appendix.

\subsection{Dataset Analysis}
\label{subsec:pipeline_analysis}

\begin{table*}[t]
    \centering
    \caption{The average change in unique nouns and verbs relative to the ground truth, as well as word count and length for the different dimensions of ActivityNet10k, vs. ActivityNet Captions~\cite{krishna2017densecaptioning}.
    }
    \vspace{-0.5em}
    \resizebox{1.\linewidth}{!}{
    \begin{tabular}{@{}p{70pt}cccccccccc@{}}
    \toprule
                  &      & \multicolumn{3}{c}{Summarization} & \multicolumn{3}{c}{Simplification} & \multicolumn{3}{c@{}}{Summarization and Simplification}        \\
    \cmidrule(l){3-5} 
    \cmidrule(l){6-8}
    \cmidrule(l){9-11}
    Metric & Source & Short & Medium & Full Length & Elementary & Intermediate & University & S and E & S and I & S and U \\
    \midrule
    $\Delta$ Nouns & & -5.37 & -1.69 & 0.06 & -1.16 & 0.97 & 3.84 & -6.20 & -5.83 & -5.41 \\
    $\Delta$ Verbs & & -5.01 & -1.97 & -0.77 & -0.65 & 0.90 & 1.95 & -5.19 & -5.02 & -4.84 \\
    Word Count & 49.77 & 8.77 & 29.29 & 37.39 & 43.54 & 48.31 & 56.13 & 8.50 & 9.28 & 10.75 \\
    Word Length & 5.09 & 6.10 & 5.40 & 5.51 & 4.97 & 5.49 & 5.97 & 5.27 & 5.75 & 6.10 \\
    \bottomrule
    \end{tabular}
    }
    \label{tab:automatic_statistics}
\end{table*}

\begin{table*}[ht]
    \begin{minipage}{0.31\linewidth}
        \caption{\textbf{Meaning Preservation} results. For each item, we present the annotators with a paragraph
        and three synthetic captions: one generated from the paragraph, one from a neighbor, 
        and one at random. We show how often each caption is judged as a match to the paragraph.
        }
        \vspace{-0.5em}
        \label{tab:meaning_preservation}
        \resizebox{1.\textwidth}{!}{
            \setlength{\tabcolsep}{8pt}
            \begin{tabular}{@{}l c c c@{}} 
                \toprule
                 & Different & Unsure & Matches \\
                \midrule
                Actual Match  & 0\% & 4\% & 96\% \\
                Neighbor  & 20\% & 28\% & 52\% \\
                Random & 100\% & 0\% & 0\% \\
                \bottomrule
            \end{tabular}
        }
        \end{minipage} 
    \hfill
    \begin{minipage}{0.31\linewidth}
        \caption{\textbf{Simplification Validation} results. For each paragraph we ask the annotators to rank 
        the three synthetic simplification captions from simplest to most complex.
        The majority results show that actual complexities correlate well with the intended simplification.}
        \vspace{-0.5em}
        \label{tab:simplification_validation}
        \resizebox{1.\textwidth}{!}{
            \setlength{\tabcolsep}{8pt}
            \begin{tabular}{@{}l c c c@{}} 
                \toprule
                 & Simplest & Middle & Most Complex \\
                \midrule
                Elementary  & 84\% & 16\% & 0\% \\
                Intermediate  & 16\% & 84\% & 0\% \\
                University & 0\% & 0\% & 100\% \\
                \bottomrule
            \end{tabular}
        }
        \end{minipage} 
    \hfill
    \begin{minipage}{0.31\linewidth}
        \caption{\textbf{Hallucination Prevalence} results. We treat each potentially hallucinated word 
        in the generated caption as an item, and show results over all votes, as well as the majority label for each item.
        This suggests most potential hallucinations are actually consistent with the source caption.}
        \vspace{-0.5em}
        \label{tab:hallucination_prevalence}
        \resizebox{1.\textwidth}{!}{
            \setlength{\tabcolsep}{8pt}
            \begin{tabular}{@{}l c c c@{}} 
                \toprule
                 & Different & Unsure & Matches \\
                \midrule
                Total  & 24.75\% & 8.08\% & 67.17\% \\
                Majority (per-word)  & 18.18\% & 9.09\% & 72.73\% \\
                \bottomrule
            \end{tabular}
        }
        \end{minipage} 
    \end{table*}


\begin{table}[ht]
\centering
    \begin{minipage}{1.\linewidth}
    \centering
    \caption{\textbf{Unanimous annotator agreement}, or the portion of items per 
    section for which \textit{all} annotators give the same label.}
    \label{tab:annotator_agreement}
    \vspace{-0.5em}
    \resizebox{0.8\textwidth}{!}{
        \setlength{\tabcolsep}{8pt}
        \begin{tabular}{@{}l c c c@{}}
            \toprule
             & \makecell{Meaning\\ Preservation} & \makecell{Simplification\\ Validation} & \makecell{Hallucination\\ Prevalence} \\
            \midrule
            Actual  & 71.67\% & 64.00\% & 55.00\% \\
            \bottomrule
        \end{tabular}
    }
    \end{minipage}
    \vspace{-0.5em}
\end{table}

We provide some fine-grained statistical measures to examine the nature of our generated data (\textbf{Automatic Analysis}). 
We also perform a study on a sample of our data using human annotators to further validate the claims regarding our data and ensure that it is free from undesirable artifacts (\textbf{Annotator Analysis}).

\textbf{Automatic Analysis.}
For the sake of brevity we focus on ActivityNet, with metrics
for other datasets provided in the appendix.
From Table~\ref{tab:automatic_statistics}, note that summarization and elementary level simplification tend to remove nouns and verbs \footnote{Extracted using \url{https://spacy.io} part-of-speech tagging},
while higher reading levels tend to add
nouns and verbs.
Also note the word counts, where summarization, as expected, reduces the average number of words,
while simplification to elementary level reduces average word length.

\textbf{Annotator Analysis.}
To validate the fidelity and utility of our captions we 
recruit 15 human annotators to examine our captions
in an IRB approved study.
We design a survey that consists of 3 sections, according to the properties 
of the data we wish to examine.
In the first section, we analyze whether the LLM-generated captions 
preserve original meaning.
In the second section, we ensure that when the LLM performs the simplification 
in a manner that is meaningful to humans.
In the third section, we verify the extent to which hallucinations 
occur in the LLM-generated captions.
Each section has 5 questions.
We divide our annotators into groups of 3, to allow for analysis of inter-annotator
agreement, and thus distribute 5 versions of the survey, covering a sample 
of 75 videos from the validation set of ActivityNet.
Next, we provide more detail regarding the design of the survey and results for each section.
As evidence of the survey's validity, we show inter-annotator agreement in 
Table~\ref{tab:annotator_agreement}.
For full details, please see the appendix.

\textbf{Meaning Preservation.} 
For each question in this section, we randomly sample one real caption, 
and assign it to be the ``ground truth'' caption. 
We then sample 3 generated captions -- one generated from the 
``ground truth'' caption, one generated from the ``ground truth'' 
caption's nearest neighbor caption~\footnote{From cosine similarity 
between captions using OpenAI's \texttt{text-embedding-ada-002}.}, and one generated from a random unrelated caption.
We ask the annotator to determine, for each of the 3 generated captions, 
whether they believe it describes the same video as the ground truth caption.
Table~\ref{tab:meaning_preservation} shows that the synthetic caption is very 
consistently judged to be from the same video as its source caption, unlike neighbor and random captions.

\textbf{Simplification Validation.} 
For each question in this section, we randomly sample one real caption and show the annotator the ``elementary'', 
``intermediate'', and ``university'' captions generated by the LLM.
Then we ask the annotator to rank them from most to least complex. 
In Table~\ref{tab:simplification_validation}, we find very little ambiguity in the simplification rankings.
Annotators consistently judge ``university'' to be the most complex, and 
``elementary'' to be the least complex.

\textbf{Hallucination Prevalence.} 
For each question in this section, we sample some real caption and 
one of its generated captions, either the full length summary, or one of the 
full length simplification captions.
We then use spaCy part-of-speech tagging to extract the nouns and verbs 
which appear in the generated caption but not the original. 
We then present both captions to the annotator, and for up to 3 of the 
potentially hallucinated words, we ask whether or not they change the 
meaning of the original caption. 
We find, in Table~\ref{tab:hallucination_prevalence}, that of the new words for the generated captions, annotators tend to 
judge that they typically correspond to entities and actions that are already 
depicted in the source captions. 
This, along with the results from the first section of the study,
suggest that the prevalence and  impact of potential hallucination is 
quite limited. 

\section{Benchmark Results}
\label{sec:benchmark}

\begin{table*}[t]
    \centering
    \caption{Text to Video retrieval performance for our benchmark on ActivityNet. First
    we reproduce results for standard paragraph to video retrieval. Then, we give the average
    performance on short 10k Words captions, long 10k Words captions, and partial captions.
    We use the standard recall at top-1 metric (R@1) as well as the average of recall at top-1/5/10 (Avg. Recall). 
    We explain how we aggregate 10k Words captions as Full, Short, Long, and Partial in Section~\ref{sec:benchmark}.
    We explain the finetuning in Section~\ref{sec:improving_performance}. 
    \vspace{-0.5em}
    }
    \resizebox{1.\linewidth}{!}{
    \begin{tabular}{@{} cl cc cc cc cc cc cc cc @{}}
    \toprule
    \multicolumn{2}{c}{\multirow{3}{*}{Model}} & \multicolumn{6}{c}{{\bf Standard}} &\multicolumn{8}{c}{{\bf 10k Words}} \\
    \cmidrule(l){3-8}
    \cmidrule(l){9-16}
                              && \multicolumn{2}{c}{ANet Full}            & \multicolumn{2}{c}{QuerYD Full}          & \multicolumn{2}{c}{LF-VILA Full} & \multicolumn{2}{c}{ANet10k All}         & \multicolumn{2}{c}{ANet10k Short} & \multicolumn{2}{c}{ANet10k Long} & \multicolumn{2}{c}{ANet10k Partial}     \\
    \cmidrule(l){3-4}
    \cmidrule(l){5-6}
    \cmidrule(l){7-8}
    \cmidrule(l){9-10}
    \cmidrule(l){11-12}
    \cmidrule(l){13-14}
    \cmidrule(l){15-16}
                        && R@1             & Avg. R             & R@1             & Avg. R             & R@1             & Avg. R             & R@1          & Avg. R     & R@1          & Avg. R    & R@1          & Avg. R  & R@1 & Avg. R        \\ 
    \midrule
    \multirow{4}{*}{zero-shot}& VideoCLIP        & 6.3 & 16.2 & 7.4 & 15.9 & 5.1 & 11.1 & 5.3 & 13.9 & 4.0 & 10.9 & 6.2 & 16.3 & 6.6 & 16.3 \\
    & Frozen                                     & 14.0 & 32.4 & 13.7 & 27.5 & 26.1 & 43.4 & 11.4 & 27.4 & 8.7 & 22.8 & 14.2 & 32.0 & 11.0 & 27.3 \\
    & COSA                                       & 34.2 & 56.2 & 34.4 & 49.6 & \textbf{66.8} & \textbf{78.4} & 23.9 & 43.4 & 16.2 & 33.9 & 31.7 & 52.9 & 23.7 & 43.4 \\
    & InternVideo                                & \textbf{47.9} & \textbf{68.2} & \textbf{50.1} & \textbf{63.6} & 49.5 & 63.5 & \textbf{36.0} & \textbf{57.5} & \textbf{27.8} & \textbf{49.3} & \textbf{44.4} & \textbf{65.9} & \textbf{35.0} & \textbf{56.8} \\
    \midrule
    \multirow{2}{*}{finetune}&Domain            & 59.1 & 77.7 & - & - & 95.9 & 98.6 & 43.0 & 64.2 & 29.1 & 51.5 & 54.2 & 74.5 & 43.2 & 64.5 \\
    & Ours                                  & \textbf{60.1} & \textbf{78.7} & - & - & \textbf{97.4} & \textbf{99.1} & \textbf{45.8} & \textbf{67.6} & \textbf{32.9} & \textbf{57.1} & \textbf{56.5} & \textbf{76.5} & \textbf{44.3} & \textbf{65.9} \\
    \bottomrule
    \end{tabular}
    }
    \label{tab:main_results}
\end{table*}

From Section \ref{subsec:pipeline_analysis}, we conclude that the captions we generate in our paradigm are diverse and robust.
In this section we demonstrate that they are useful for benchmarking the text-to-video retrieval performance of video-language models.



\textbf{Models}
For our 10k Words benchmark, we evaluate the performance of VideoCLIP~\cite{xu2021videoclip}, Frozen~\cite{bain2022frozen}, InternVideo~\cite{Wang2022InternVideoGV}, and 
COSA~\cite{Chen2023COSACS} as a set of representative video-language models. 
For COSA we use the `itm' retrieval and for InternVideo we use the dual softmax. 
%

\textbf{Experiment Details}
We run our experiments on nodes containing between 1 and 4 GTX 2080Ti, RTX A5000, and RTX A6000 
GPUs, depending on the demands of each model.
When training, we follow the settings provided in the publicly available code of the models 
we chose. 
%

\textbf{Metrics}
We use text-to-video recall @ K (R@K) to measure performance.
Given a list of
text queries and video targets relative to a database of videos to be retrieved, R@K
measures the percentage of queries for which the ground truth target was
retrieved at rank K.
Avg. R averages R@1, R@5, and R@10.

To measure performance on the Duration axis we consider whether the partial and full captions can retrieve the full length video.
The ``\textbf{Full}'' setting measures how
often the full caption (f) retrieves the video at rank K or better, which represents performance as measured by the standard datasets.
Since we use the standard ActivityNet settings, these can be compared with numbers from other papers; however, since we use unique splits for QuerYD and LF-VILA, these numbers are not comparable.
The ``\textbf{Partial}''
setting measures how often the partial caption
(p) retrieves the same full length video.
The ``\textbf{Short}'' setting measures performance of full length video retrieved by 
short captions including short summarization (s) and simplifications of it (s+e, s+i, s+u). 
Similarly, we also report performance on the ``\textbf{Long}'' setting which include 
long summarization (l) and simplifications of it (l+e, l+i, l+u). 
The ``\textbf{All}'' setting is an average of \textbf{Partial}, \textbf{Short}, and \textbf{Long}, weighted by the number of caption types for each. 

\begin{table}[t!]
    \centering
    \caption{We provide further zero shot results (R@1) for LF-VILA10k and QuerYD10k.}
    \resizebox{0.95\linewidth}{!}{
    \begin{tabular}{@{}l cccc cccc@{}}
    \toprule                          
                              & \multicolumn{4}{c}{QuerYD10k}                        & \multicolumn{4}{c}{LF-VILA10k}                       \\
    \cmidrule(l){2-5}
    \cmidrule(l){6-9}
    Model                    & All & Short   & Long    & Partial       & All & Short   & Long   & Partial      \\ 
    \midrule
    VideoCLIP   & 6.8 & 7.0  & 6.4  & 7.8  & 4.3 & 4.1  & 4.4 & 5.0 \\
    Frozen      & 13.3 & 12.2 & 15.1 & 10.5 & 21.4 & 17.7 & 25.6 & 19.3 \\
    COSA        & 27.8 & 27.4 & 29.6 & 21.9 & \textbf{42.8} &\textbf{41.0} & 44.0 & \textbf{45.4} \\
    InternVideo & \textbf{46.8} & \textbf{44.8} & \textbf{49.2} & \textbf{45.3} & 39.9 & 35.4 & \textbf{45.2} & 37.1 \\
    \bottomrule
    \end{tabular}
    }
    \label{tab:other_dataset_results}
\end{table}

We provide the zero-shot benchmarking results in Table~\ref{tab:main_results}, with remaining results
on the LF-VILA and QuerYD datasets in Table~\ref{tab:other_dataset_results}.
The methods that perform best for Full paragraphs also tend to perform best for the Long, 
Short, and Partial captions. 
Notably, there is only a minor gap in retrieval performance between the Full and Long captions, 
with a larger difference between Full and Partial captions, and a significant drop for the 
Short captions.
We also see that COSA seems to be by far the least robust to 10k Words data, with the largest relative changes in performance.
By contrast, VideoCLIP and especially Frozen are often benefited by the 10k Words data, particularly when the axis is rewording the caption (e, i, u) rather than removing information from it (s, p).

\section{Improving Performance}
\label{sec:improving_performance}

In this section we present baseline results where we finetune pre-trained models to retrieve videos from text.
We then explore two main ways to leverage our data to improve these results.
The first is at training time -- with no extra cost in terms of parameters, iterations, or FLOPS, we can train with synthetic captions to improve retrieval results.
The second is at inference time -- we can leverage the 10k prompts as a form of query expansion, and aggregate the retrievals across equivalent 10k captions.

\subsection{Training-time Improvements}
\label{subsec:training}

We propose an approach for leveraging our 10k data that is lightweight and flexible, allowing us to perform finetuning both COSA and InternVideo.
We sample a batch of videos with corresponding captions and apply a loss that
pushes matching video and caption embeddings closer together.
To encourage the model to associate all of the descriptions for a video with that video we also include the synthetic captions for a video during training.

Specifically, for every video we sample (i) the ground truth paragraph, and (ii) a random 10k Words caption. 
We mix the two sets of captions, taking one caption per video, to yield our primary text features, $f_t$, ensuring that a fixed percentage (set by a mixing ratio, $\eta$) are 10k Words captions, and 
the rest are ground truth.
Using these primary text features, we compute standard bi-directional contrastive loss with the 
video features as in COSA~\cite{Chen2023COSACS} and InternVideo~\cite{Wang2022InternVideoGV}.
The advantage of such a simple formulation, is that it is easily reusable across many SOTA video-language models, since most leverage some video-text contrastive loss.
This allows us to apply it to both COSA (Table~\ref{tab:main_results}) and InternVideo (Table~\ref{tab:internvideo_finetune_results}).

We use 2 settings in this paper: ``Domain Finetune'' which is just the default setting of whichever model (COSA or InternVideo) we are finetuning (with no synthetic captions), and ``Ours'' where we set $\eta=0.75$. We set these values after some ablations, although ultimately these ablations (see Appendix) suggest that models are not sensitive to the exact $\eta$ so long as it is not extremely high or low.

\begin{table}[h]
        \centering
        \caption{ActivityNet InternVideo finetuning. 
        Ours is best.}
        \label{tab:internvideo_finetune_results}
        \resizebox{1.\linewidth}{!}{
        \begin{tabular}{@{} l cc cc cc cc @{}}
        \toprule
        \multirow{2}{*}[-0.25em]{\shortstack{Finetune \\Method}}                  & \multicolumn{2}{c}{All}                       & \multicolumn{2}{c}{Partial}                 & \multicolumn{2}{c}{Short}      & \multicolumn{2}{c}{Long}      \\
        \cmidrule(l){2-3}
        \cmidrule(l){4-5}
        \cmidrule(l){6-7}
        \cmidrule(l){8-9}
                           & R@1          & Avg. R                  & R@1          & Avg. R                  & R@1          & Avg. R              & R@1          & Avg. R      \\ 
        \midrule
        Domain                   & 41.4 & 62.7 & 52.1 & 72.4 & 29.1 & 51.2 & 48.5 & 69.7                \\
        Ours                      & \textbf{42.2} & \textbf{63.6} & \textbf{52.6} & \textbf{72.8} & \textbf{30.0} & \textbf{52.6} & \textbf{49.1} & \textbf{70.5}                \\
        \bottomrule
        \end{tabular}
        }
\end{table}

\begin{figure}[ht]
    \centering
    \includegraphics[width=1.0\linewidth]{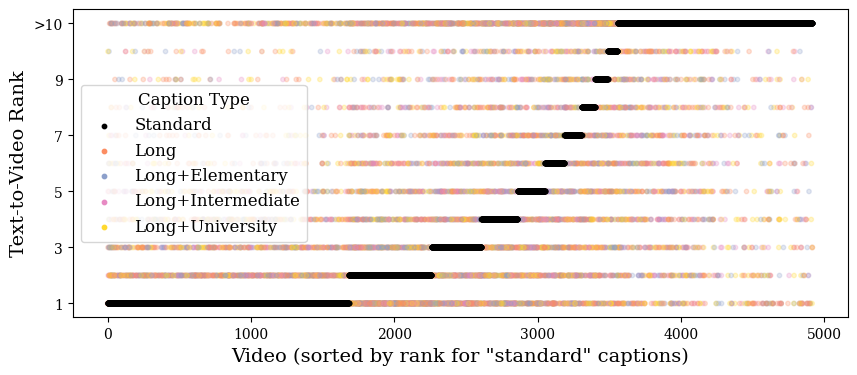}
    \caption{
        We plot standard caption retrieval results for each item in ActivityNet, sorted by rank.
        We also plot the retrieval for a few synthetic caption types, sorted by standard caption retrieval rank.
        For many samples, synthetic captions yield superior retrievals.
    }
    \label{fig:overlaps}
\end{figure}

Table~\ref{tab:main_results} shows the results for COSA finetuning on ActivityNet (``\textbf{Domain}'').
We observe that finetuning by sampling from 10k Words data (``\textbf{Ours}'') yields considerable improvements for retrieving 
with 10k Words captions.
We show that these findings hold when we adapt our data sampling and losses for another 
state-of-the-art model, InternVideo, in Table~\ref{tab:internvideo_finetune_results}, as well as when finetuning on other datasets, 
such as LF-VILA (see appendix).
While the improvements on InternVideo are less dramatic (perhaps because it is closer to the best-case performance, see Section~\ref{sec:understanding_failures}, they are still nontrivial, and give evidence for the effectiveness of our data.
We find that LF-VILA10k responds more to finetuning, perhaps because it is open-domain and lacks the fine-grained difficulties of ActivityNet.

\begin{table*}[t]
\centering
    \begin{minipage}{0.3\linewidth}
        \centering
        \caption{ANet query ensemble retrieval results.
        We compare standard retrieval on the zero-shot, domain-finetuned, and 10k-finetuned COSA models to retrieval with an ensemble of synthetic captions. The ensemble is very effective for zero-shot.}
        \label{tab:ensemble_results}
        \vspace{-0.5em}
        \resizebox{1.\linewidth}{!}{
        \begin{tabular}{@{} ll cc @{}}
        \toprule
        Finetuning & Inference & R@1 & Avg. R \\
        \midrule
        \multirow{2}{*}{Zero-shot} & Standard & 34.2 & 56.2               \\
         & Ensemble & \textbf{37.6} & \textbf{58.6} \\
        \midrule
        \multirow{2}{*}{Domain} & Standard & 59.1 & 77.7               \\
         & Ensemble & \textbf{59.2} & \textbf{77.9} \\
        \midrule
        \multirow{2}{*}{Our} & Standard & 60.1 & 78.7                \\
         & Ensemble & \textbf{60.2} & \textbf{78.8 }\\
        \bottomrule
        \end{tabular}
        }
    \end{minipage} 
    \hfill
    \begin{minipage}{0.675\linewidth} 
        \centering
        \caption{We show some example failures, where the target video is retrieved outside the top 10 despite the short description's uniqueness and specificity. For each failure, we give the short caption (S), the GT caption of the correct video (C), and an incorrec top 10 retrieval (I).}
        \label{tab:bad_retrieval_examples}
        \vspace{-0.5em}
        \resizebox{1.\linewidth}{!}{
        \begin{tabular}{@{} ll @{}}
        \toprule
        Type & Caption \\
        \midrule
        S & Man throws ball, goalkeeper blocks. \\
        C & The man threw... As the players throw the ball to the goal, the goalkeeper blocked \\
        & the ball. The players swim... \\
        I & Two teams play... soccer. A goal is scored... with a bicycle kick... \\
        \midrule
        S & Two girls play dress up, laughing and drying their faces on a towel. \\
        C & Two girls are playing dress ... laughing. One girl dries her face on a towel... second \\
        & girl dries her face... \\
        I & A woman is sitting in a chair. Another woman... starts clipping and filing the other \\
        & woman's nails. \\

        \bottomrule
        \end{tabular}
        }
    \end{minipage} 
\end{table*}

\subsection{Inference-time Improvements}
\label{subsec:inference_improvements}

The 10k data can improve performance without the need for training at all.
It can be used as a form of query expansion to also improve performance at inference time.
We already know from Section~\ref{sec:benchmark} that overall, standard captions retrieve videos with more recall than 10k captions.
However,in Figure~\ref{fig:overlaps}, we find that the correct retrievals using standard captions are not a superset of the correct retrievals when using synthetic 10k captions.
That is, there exist samples where while the standard caption does not retrieve the video well, some 10k caption does.
So, we hypothesize that if we aggregate the predictions by attempting the retrieval with both synthetic and standard captions for each sample (instead of only the standard caption), we can improve the quality of the retrieved results.

Our aggregation method is simple. 
First, we determine which types of 10k captions we will use (see Appendix).
We then compute the standard text-video similarity matrix, as well as a separate text-video similarity matrix for each type of 10k caption we choose.
We then add these together, giving $50\%$ weight to the standard text-video matrix, and equal weight to the remaining matrices.
We report results for performing this sort of query expansion ensemble with COSA zero-shot, domain finetuning, and ours in Table~\ref{tab:ensemble_results}.

\section{Understanding Failures}
\label{sec:understanding_failures}

To gather insight into whether our dataset is inherently difficult or whether our models should be able to perform better on the 10k Words benchmark, we analyze why models perform much more poorly for short captions than for long captions. 
The models we finetune are not pretrained on ActivityNet Captions, so there is clearly a domain gap between the training and testing distributions.
Our 10k datasets add an additional domain gap for each axis of augmentation.
If this domain gap were the primary difficulty introduced by 10k data, then finetuning on the data as we do in our approach would result in similar performance across the different types of 10k descriptions.
Instead we see that performance on short captions is much lower than performance on long captions.
We investigate this by focusing on two questions.
(i) How does the information in a caption affect model performance?
(ii) Is each short caption specific enough to uniquely match the corresponding video?

\begin{figure}[h]
    \centering
    \includegraphics[width=0.95\linewidth]{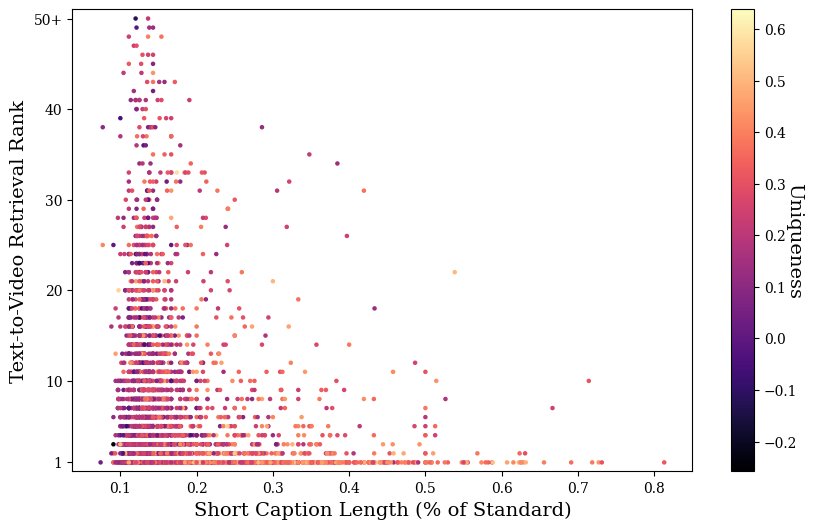}
    \caption{
        We measure the length and retrieval uniqueness for short caption retrieval, and find that the highest ranks correlate with captions that have lost their unique information.
    }
    \label{fig:rank_vs_length_w_unique}
    \end{figure}

\begin{figure}[h]
    \centering
    \includegraphics[width=1.0\linewidth]{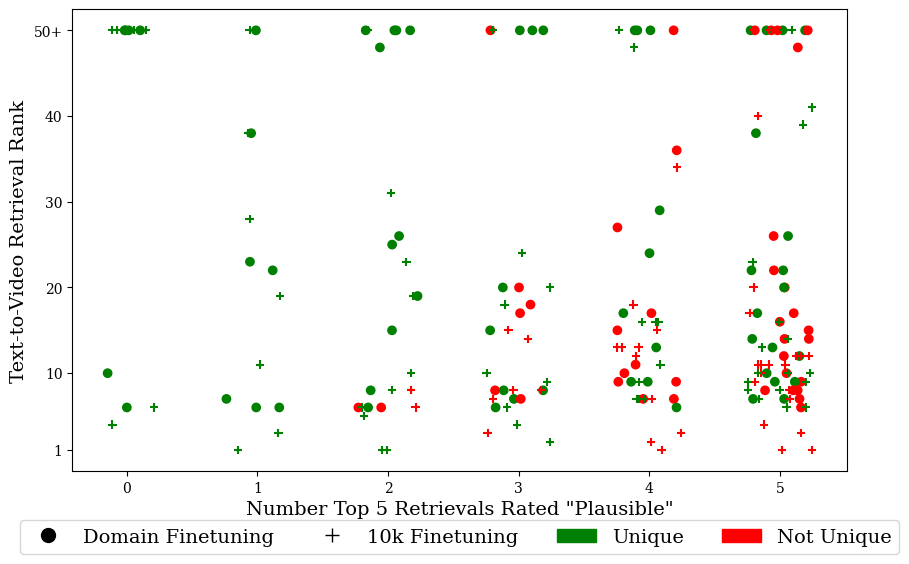}
    \vspace{-2.0em}
    \caption{
        We measure uniqueness and plausibility for short captions with bad retrievals. We find that most difficult samples tend to be non-unique and have many plausible correct retrievals.
    }
    \label{fig:bad_retrievals_analysis}
    \vspace{-1.0em}
\end{figure}

\textbf{Information Loss.}
We experiment with three different ways to practically measure the information content of a caption, including an approach based on counting the number of entities relative to the full caption and an approach based on embedding similarities.
In Figure~\ref{fig:rank_vs_length_w_unique} we compute information as length of the short caption (measured by raw number of words) divided by the length of the standard caption, which we find provides very similar results to the other approaches (see Appendix) and is simpler.
Our results in Figure~\ref{fig:rank_vs_length_w_unique} show that most bad retrievals (high rank) occur when information amount is relatively low, agreeing with the coarse analysis from Table~\ref{tab:main_results}.
To understand this further we also report relative word counts conditioned on correctness.
For samples where both short and standard captions retrieve the matching video at top-1, the short caption has on average 19.1\% as many words as the standard caption.
When the standard retrieval works and the short fails to retrieve at top-1, this is 15.6\%.
Clearly, when the synthetic short caption is closer to the length of the original, the retrieval tends to be easier.
However, for some samples the retrieval succeeds when the caption is relatively short, indicating there are other relevant factors.

\textbf{Uniqueness (Automatic).}
One such factor, uniqueness is the extent to which the information in the short caption overlaps the matching video in ways it does not overlap other videos. 
We investigate this automatically by measuring how similar sentence embeddings of short captions are to the standard captions of the top-5 actually retrieved videos (not including the matching video). 
The uniqueness color scale in Figure~\ref{fig:rank_vs_length_w_unique} is darker when these embeddings are close (less unique) and lighter when they are far apart (more unique).
Less unique instances tend to appear at higher (worse) ranks.
See the appendix for more details on the calculation.
We validate this by comparing the average uniqueness for short captions where the model retrieves the correct video at top-1, 0.304, to the uniqueness of the short captions for the samples where the model fails, 0.239.
Longer captions more uniquely identify the videos they are supposed to retrieve.
More significantly, the figures show many instances where the caption is short but relatively light in color (more unique), suggesting that many of the shorter captions should be able to recall
the correct video.

\textbf{Uniqueness (Manual).}
We also investigate the uniqueness factor manually.
Two authors annotate the top 5 retrievals for 100 failed retrievals (using short captions, where the ground truth video is not in the top 5 for the Domain Finetuned COSA model).
Specifically, they compare the short text caption (query), the standard paragraph caption it came from, and the standard paragraph captions of the top 5 retrieved videos.
They decide whether the short caption could reasonably describe the same video as the paragraph caption for each of the 5 retrieved videos (whether a retrieval is ``plausible'').
They then indicate whether the ground truth matching paragraph is a closer match to the short caption than all the top 5 retrieved videos' paragraphs (whether the short caption is ``unique'').
They then resolve differences for all 500 ``plausible'' and 100 ``unique'' labels.

We plot the results of this analysis in Figure~\ref{fig:bad_retrievals_analysis}, with the x axis indicating how many top 5 retrievals were plausible and the y axis indicating the actual rank of the correct video.
Here we see that the rank of the ground truth video is higher when more plausible alternatives are retrieved; the majority of the ``bad'' retrievals are ambiguous.
Surprisingly, sometimes the retrieval is performed successfully even in the cases where the humans considered the text quite ambiguous (5 ``plausible'' and not ``unique'').
Most significantly, there exist failed retrievals for short captions which the human labelers did not consider ambiguous, where none of the top-5 retrievals seem ``plausible'' and the caption is considered a ``unique'' match.
Models could improve their performance on these captions.
Table~\ref{tab:bad_retrieval_examples} shows some examples of these.
Models struggle with fine-grained domains, like the description of a game of water polo (where players ``throw'' balls), for which the model seems to tend to retrieve videos of soccer (where players ``kick'' balls) due to the presence of a goalkeeper.
Sometimes, it also seems the model is ignoring other key parts of the short caption, such as the presence of the towel in the second example.

\section{Related Work}
\label{sec:related_work}

\textbf{Video-Language Models.}
Video-language models build on image-based vision-language models.
These approaches are typically pre-trained in a manner inspired by 
CLIP~\cite{radford2021learning} and ALIGN~\cite{jia2021scaling},
using sets of video and text pairs with varying levels of 
noise~\cite{miech2019howto100m,bain2022frozen,zhang2023visionlanguage}.
These models are typically pretrained on some task or set of tasks,
and then used on downstream tasks either in a zero-shot manner or after finetuning.
Early approaches use pre-computed features to represent 
videos~\cite{zhang2018s3d}.
Typical models learn a shared embedding space between the videos and 
text~\cite{xu2021videoclip,alayrac2020selfsupervised,gabeur2020multimodal,ging2020coot,miech2019howto100m,miech2020endtoend,huang2021multilingual}.
Others process concatenated video and text inputs with cross-modal 
encoders~\cite{sun2019videobert,li2019unicodervl,su2020vlbert,chen2020uniter,du2023video,zhu2020actbert,xu2021vlm}.
Some even use still images or average frame embeddings and achieve quite strong performance~\cite{buch2022revisiting,lei2022revealing,bain2022cliphitchhikers}.
However, as computational resources have scaled, so have the methods, and many current approaches 
learn to compute features from raw video~\cite{bain2022frozen,Wang2022InternVideoGV,Chen2023COSACS,li2023unmasked}.
Along with this trend models are branching out from contrastive learning to incorporate other
learning tasks as well, even generative objectives including 
captioning~\cite{Wang2022InternVideoGV,Chen2023COSACS,xu2023mplug2,chen2023vast,he2023vlab,chen2023valor,yan2023videococa,kuo2023mammut}.
In this paper we focus on the video retrieval task, and show results using 
VideoCLIP~\cite{xu2021videoclip}, Frozen~\cite{bain2022frozen}, COSA~\cite{Chen2023COSACS}, 
and InternVideo~\cite{Wang2022InternVideoGV} as a representative set of models.

\textbf{Long Video Understanding.}
Videos in the computer vision literature tend to be short -- 
the average length of videos in tentpole 
datasets~\cite{rohrbach2016movie,xu2016msr-vtt,goyal2017something,chen-dolan-2011-collecting,hendricks17iccv}  
is under 30 seconds.
Over the years, some have introduced datasets consisting of longer 
videos~\cite{ghanem2018activitynet,bain2020condensed,zhou2017automatic,oncescu2021queryd,sun2023longform}.
With the introduction of these datasets, larger GPUs, and advancements in vision-text modeling, 
many researchers have begun proposing methods that either address long video 
as a first-class 
interest~\cite{bain2022cliphitchhikers,sun2023longform,ren2023testa}, 
or at least, are flexible for both long and short 
videos~\cite{li2023unmasked,chen2023valor,chen2023vast,Wang2022InternVideoGV,Chen2023COSACS}.
In this paper we focus on long video in terms of retrieval, 
and we propose a method for data expansion to enable better training and understanding of long video models.
This is reminiscent of \cite{wang2022multiquery}; however, we synthesize novel captions, 
and restrict our training and analysis to the single query retrieval setting.

\textbf{Text Summarization.} 
Our work bears some resemblance to efforts in the areas of controllable text summarization and simplification.
For these tasks, control tokens dictate how a model simplifies or summarizes text while preserving its meaning.
The definition of these tokens is a key differentiator between papers~\cite{agrawal2023control} --
they can be 
user-defined~\cite{yanamoto-etal-2022-controllable,nishihara-etal-2019-controllable,agrawal-etal-2021-non,agrawal-carpuat-2022-imitation,zetsu-etal-2022-lexically}, 
or optimized over some 
data~\cite{martin-etal-2022-muss,maddela-etal-2021-controllable,qiao-etal-2022-psycho,kew-ebling-2022-target,sheang-saggion-2021-controllable}.
Additionally, these pipelines often involve a human-in-the-loop at inference time 
to give keywords~\cite{he2020ctrlsum} and can require heavy labeling~\cite{xu-etal-2015-problems,agrawal2023control}.
We opt to use our LLM approach to avoid a reliance on control token 
definition, human-in-the-loop, or specially annotated data.

\section{Conclusion}
\label{sec:conclusion}
We showed how the data for the video retrieval problem can be expanded to
capture a greater variety ways someone might describe a video, instantiated with ActivityNet10k, QuerYD10k, and LF-VILA10k.
We showed that SOTA models fail to generalize well to all potentially
valid descriptions, and propose fine-tuning and inference-time approaches to mitigate these shortcomings.
We also distilled insights on how some SOTA models struggle with short captions.
We hope future work further explores the complete spectrum of language that can describe video content.

\noindent\textbf{Acknowledgements.} This work was partially supported by NSF CAREER Award (\#2238769) to Abhinav Shrivastava. The U.S. Government is authorized to reproduce and distribute reprints for Governmental purposes notwithstanding any copyright annotation thereon. The views and conclusions contained herein are those of the authors and should not be interpreted as necessarily representing the official policies or endorsements, either expressed or implied, of NSF or the U.S. Government.


\clearpage
\maketitlesupplementary

\section{GPT-3.5 Details}

\subsection{Prompts and Costs}

We share prompts for summarization, simplification, and the combination of the two (joint).
In the main paper, summarization is denoted as s, m, l depending on length, where s has 1 word and m has 4 words for every 7 words in l. 
Simplification is denoted by l+e, l+i, l+u. 
Joint is s+e, s+i, s+u.

We reduce the cost in terms of input token counts by batching our inputs.
For example, we are generating 3 different summarizations per paragraph, but the source paragraph is the same in all 3 cases.
So, instead of passing the input once for each level of summarization (3 times total), we pass the input once, and ask for all summarizations to be present in the output, reducing our input tokens by a factor of 3.
We do the same for simplification and joint.
So, if we want to generate summarization, simplification, and joint captions for a given ground truth caption, we must make 3 calls to the API (or, if hosted locally, one would have 3 forward passes).
Remarkably, the model did not generate a malformed response a single time; in every case, we received each of the 3 requested outputs, properly tagged.
It is worth mentioning these could possibly all be batched for a single pass, although at the time of preparing the dataset, the model was less robust under such conditions.
If using our strategy for query expansion, discussed in Section~\ref{subsec:inference_improvements}, one would ideally batch all desired axes for a single pass, for the sake of speed.

The resulting costs can be computed in terms of tokens. 
The summarization prompt is approximately 180 tokens, not including the paragraph. 
For the 14,926 ActivityNet videos we consider, whose captions are an average of 49.8 words per caption, this means we submitted approximately 3.5 million input tokens for the 3 levels of summarization.
Input tokens for the other two axes can be computed similarly.
If using certain proprietary models, one must also consider the cost for output tokens, which can be estimated based on the length of the input paragraph compared to the word counts we provide for each dimension in Table~\ref{tab:automatic_statistics}.
So, our final prompts are as follows for summarization, simplification, and joint. Note the use of ``primary school'' to generate our ``elementary'' level captions, and ``secondary school'' to generate ``intermediate'' captions.

\begin{description}
    \item[Summarization] You are a helpful writing assistant, with a speciality in summarizing text-based scene descriptions. You will be asked to write 3 summaries of the scene described in the following paragraph, indicated by PARAGRAPH. Do not modify the indicated order of events. Prioritize visual details. Do not hallucinate. Do not describe objects or events that do not appear in the original paragraph. \\
    PARAGRAPH: \textit{$\langle$ORIGINAL PARAGRAPH$\rangle$}. \\
    Label this summary as SUMMARY\_1. For this summary, please write 10 words which summarize the scene described by the PARAGRAPH. Do not use more or less than 10 words. Without using more than 10 words, write complete sentences. \\
    Label this summary as SUMMARY\_4. For this summary, please write 40 words which summarize the scene described by the PARAGRAPH. Do not use more or less than 40 words. Without using more than 40 words, write complete sentences. \\
    Label this summary as SUMMARY\_7. For this summary, please write 70 words which summarize the scene described by the PARAGRAPH. Do not use more or less than 70 words. Without using more than 70 words, write complete sentences. \\
    \item[Simplification] You are a helpful writing assistant, with a speciality in simplifying and rewriting descriptions for different age groups and reading levels. You will be asked to write 3 versions of the scene described in the following paragraph, indicated by PARAGRAPH. Do not modify the indicated order of events. Prioritize visual details. Do not hallucinate. Do not describe objects or events that do not appear in the original paragraph. \\
    PARAGRAPH: \textit{$\langle$ORIGINAL PARAGRAPH$\rangle$}. \\ 
    Label this version as VERSION\_primary\_school. For this version, rewrite the PARAGRAPH with 70 words to make it suitable for a primary school reading level. \\ 
    Label this version as VERSION\_secondary\_school. For this version, rewrite the PARAGRAPH with 70 words to make it suitable for a secondary school reading level. \\ 
    Label this version as VERSION\_university. For this version, rewrite the PARAGRAPH with 70 words to make it suitable for a university reading level. \\
    \item[Joint] You are a helpful writing assistant, with a speciality in summarizing text-based scene descriptions. You also have a speciality in simplifying and rewriting descriptions for different age groups and reading levels. You will be asked to use 10 words to write 3 summaries of the scene described in the following paragraph, indicated by PARAGRAPH. Do not modify the indicated order of events. Prioritize visual details. Do not hallucinate. Do not describe objects or events that do not appear in the original paragraph. \\
    PARAGRAPH: \textit{$\langle$ORIGINAL PARAGRAPH$\rangle$}. \\ 
    Label this version as VERSION\_primary\_school. For this version, rewrite the PARAGRAPH with 10 words to make it suitable for a primary school reading level. Do not use more or less than 10 words. Without using more than 10 words, write complete sentences. \\ 
    Label this version as VERSION\_secondary\_school. For this version, rewrite the PARAGRAPH with 10 words to make it suitable for a secondary school reading level. Do not use more or less than 10 words. Without using more than 10 words, write complete sentences. \\ 
    Label this version as VERSION\_university. For this version, rewrite the PARAGRAPH with 10 words to make it suitable for a university reading level. Do not use more or less than 10 words. Without using more than 10 words, write complete sentences.
\end{description}

\subsection{Automatic Analysis}

\begin{table*}[t]
    \centering
    \caption{Automatic dataset statistics for LF-VILA10k. We show the average change in unique nouns and verbs, as well as word count and length.}
    \vspace{-0.5em}
    \resizebox{1.\linewidth}{!}{
    \begin{tabular}{@{}p{70pt}cccccccccc@{}}
    \toprule
                  &      & \multicolumn{3}{c}{Summarization} & \multicolumn{3}{c}{Simplification} & \multicolumn{3}{c@{}}{Summarization and Simplification}        \\
    \cmidrule(l){3-5} 
    \cmidrule(l){6-8}
    \cmidrule(l){9-11}
    Metric & Source & Short & Medium & Full Length & Elementary & Intermediate & University & S and P & S and S & S and U \\
    \midrule
    $\Delta$ Nouns & & -11.77 & -4.23 & 1.49 & -1.40 & 3.75 & 11.14 & -14.35 & -13.18 & -12.36 \\
    $\Delta$ Verbs & & -2.60 & 0.90 & 4.32 & 1.96 & 7.63 & 11.71 & -3.07 & -2.34 & -1.86 \\
    Word Count & 155.40 & 36.06 & 76.30 & 105.43 & 129.52 & 136.58 & 154.13 & 28.94 & 31.85 & 34.83 \\
    Word Length & 4.66 & 5.00 & 4.96 & 5.18 & 4.79 & 5.25 & 5.74 & 4.66 & 4.90 & 5.12 \\
    \bottomrule
    \end{tabular}
    }
    \label{tab:automatic_statistics_lfvila}
\end{table*}

\begin{table*}[t]
    \centering
    \caption{Automatic dataset statistics for QuerYD10k. We show the average change in unique nouns and verbs, as well as word count and length.}
    \vspace{-0.5em}
    \resizebox{1.\linewidth}{!}{
    \begin{tabular}{@{}p{70pt}cccccccccc@{}}
    \toprule
                  &      & \multicolumn{3}{c}{Summarization} & \multicolumn{3}{c}{Simplification} & \multicolumn{3}{c@{}}{Summarization and Simplification}        \\
    \cmidrule(l){3-5} 
    \cmidrule(l){6-8}
    \cmidrule(l){9-11}
    Metric & Source & Short & Medium & Full Length & Elementary & Intermediate & University & S and P & S and S & S and U \\
    \midrule
    $\Delta$ Nouns & & -27.25 & -20.38 & -14.81 & -14.83 & -9.26 & -2.55 & -32.21 & -31.16 & -29.27 \\
    $\Delta$ Verbs & & -12.10 & -7.90 & -4.46 & -3.04 & 1.26 & 4.56 & -14.11 & -13.57 & -12.83 \\
    Word Count & 207.86 & 53.41 & 86.69 & 114.55 & 150.97 & 164.26 & 181.92 & 34.81 & 37.28 & 43.10 \\
    Word Length & 5.47 & 5.89 & 5.72 & 5.79 & 5.27 & 5.66 & 6.02 & 5.37 & 5.73 & 5.98 \\
    \bottomrule
    \end{tabular}
    }
    \label{tab:automatic_statistics_queryd}
\end{table*}

We provide LF-VILA and QuerYD to complement Table~\ref{tab:automatic_statistics} in Table~\ref{tab:automatic_statistics_lfvila} and Table~\ref{tab:automatic_statistics_queryd}, respectively.
These are conistent with the major trends for ActivityNet10k, with the notable difference that since these captions are longer, the absolute differences are larger.

\subsection{Annotator Analysis}

For our sample, we recruited 15 individuals, all of whom had at least a bachelor's degree.
Individuals spent between 10 and 20 minutes to answer the 15 questions on their assigned survey.
For an example survey, please refer to the attached material.

\section{Ablations}

\begin{table}[ht]
    \centering
    \caption{Mixing ratio ablations.}
    \resizebox{0.5\linewidth}{!}{
    \begin{tabular}{@{}c ccc@{}}
    \toprule                          
     & \multicolumn{3}{c}{ActivityNet}                   \\
    \cmidrule(l){2-4}
    $\eta$      & Full         & Short   & Long   \\ 
    \midrule
    0.0 & 59.4 & 31.9 & 55.8 \\
    0.25 & \textbf{60.1} & 33.2 & \textbf{56.6} \\
    0.5 & 59.4 & 33.3 & 56.5 \\
    0.75 & 59.9 & \textbf{33.5} & 56.2 \\
    1.0 & 59.3 & \textbf{33.5} & \textbf{56.6} \\
    \bottomrule
    \end{tabular}
    }
    \label{tab:proj_ablation}
\end{table}

We share some ablations that indicate how we choose hyperparameter values.
The most important thing is that the losses are used, and the change that causes the most different is training with $\eta = 0.0$, highlighting the importance of using 10k Words data while training.

\section{Miscellaneous}

\subsection{Hallucination Prevalence Results}

In Table~\ref{tab:hallucination_prevalence} we give results computed in two ways, as the percentage of all votes which belong to a given category (``Total'') and by determining the majority label for each word, then computing percentages (``Majority''). 
To further clarify this computation, consider the following example, with 3 voters and 3 words. 
For the first word, 2 voters select matches, 1 selects unsure.
For the second word, all 3 voters select unsure.
For the third word, 3 select different.
Since there were 3 votes for different, 4 for unsure, and 2 for matches, the percentages for total would be 33.33\%, 44.44\%, and 22.22\% respectively.
For majority, since the first was majority matches, second was majority unsure, and third was majority different, these would be 33.33\% each.

\subsection{Training-time Improvement Details}

\begin{figure*}[t]
    \centering
    \includegraphics[width=0.95\linewidth]{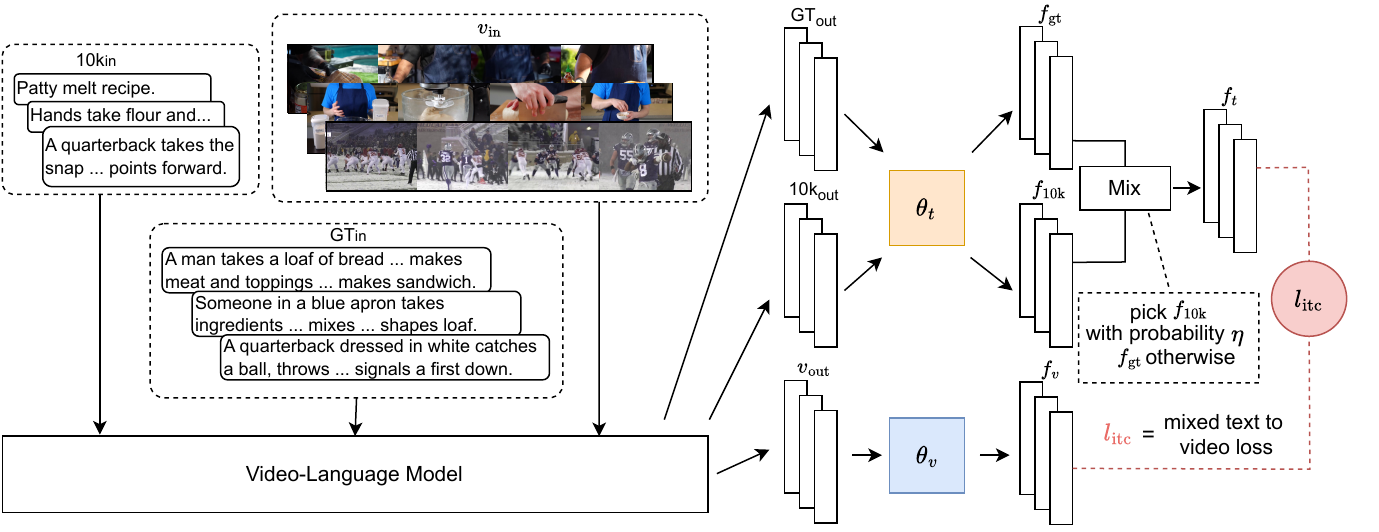}
    \caption{We perform contrastive finetuning for retrieval with video-caption pairs. \
    We propose efficient sampling of our 10k text captions for data augmentation, where we compute standard contrastive loss, but each caption is sampled \
    randomly from the 10k captions for a given video, according to a mixing ratio, $\eta$.}
    \label{fig:finetuning}
\end{figure*}

First, we show an illustration of our data sampling approach, as a visual aid, in Figure~\ref{fig:finetuning}.

\begin{figure}[h]
    \centering
    \includegraphics[width=1.0\linewidth]{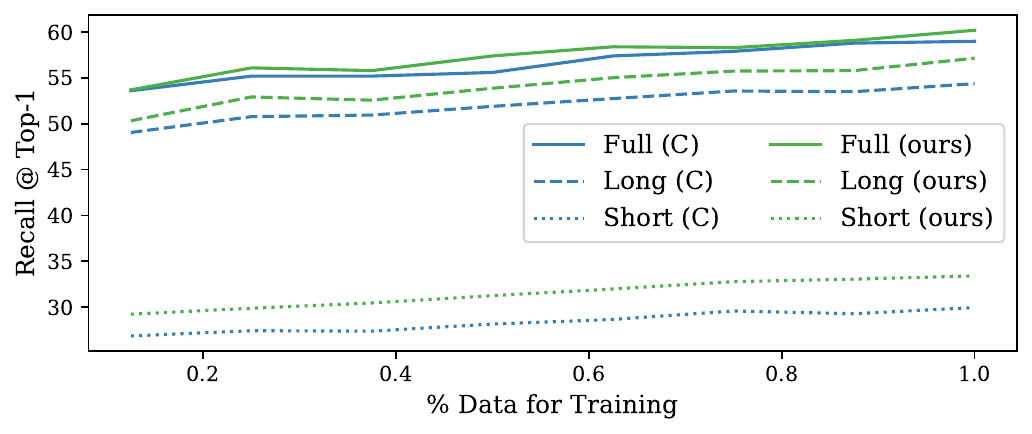}
    \caption{We measure how much our data augmentation helps in the data constrained regime,
    training only with the indicated amounts of data, and performing retrieval
    with the resulting trained models. 
    We show that finetuning COSA with 10k data (ours) is superior
    to generic COSA finetuning (C) for ActivityNet10k.}
    \label{fig:semi}
\end{figure}

Since part of our contribution is a data augmentation strategy, we also evaluate its performance by finetuning with different fractions of the original ActivityNet data in Figure~\ref{fig:semi}.
Notice that the absolute differences in recall between training with 10k data and training without remain consistent for all amounts of training data.
For training on short captions the difference is around a 3\% improvement while for long captions it is around 2\%.
By training with synthetic data, we achieve the same performance with less manually annotated data.

\begin{table}[h]
\centering
        \caption{LFVILA COSA finetuning. \
        Results improve with 10k finetuning.} 
        \label{tab:lfvila_finetune_results}
        \resizebox{1.\linewidth}{!}{
        \begin{tabular}{@{}l cc cc cc cc@{}}
        \toprule
        \multirow{2}{*}[-0.25em]{\shortstack{Finetune \\Method}}                 & \multicolumn{2}{c}{All} & \multicolumn{2}{c}{Short}          & \multicolumn{2}{c}{Long}      & \multicolumn{2}{c}{Partial}     \\
        \cmidrule(l){2-3}
        \cmidrule(l){4-5}
        \cmidrule(l){6-7}
        \cmidrule(l){8-9}
                           & R@1          & Avg. R   & R@1          & Avg. R                  & R@1          & Avg. R              & R@1          & Avg. R      \\ 
        \midrule
        Domain                   & 77.3 & 86.9 & 65.2 & 78.4 & 90.2 & 95.9 & 73.8 & 84.9                \\
        Ours                      & \textbf{85.2} & \textbf{92.6} & \textbf{78.2} & \textbf{89.2} & \textbf{95.3} & \textbf{98.2} & \textbf{73.0} & \textbf{83.9}                \\
        \bottomrule
        \end{tabular}
        }
\end{table}

We also show that our findings hold when finetuning on other datasets, such as LF-VILA (Table~\ref{tab:lfvila_finetune_results}).

\subsection{Inference-time Improvement Details}
For our ensembles in Section~\ref{subsec:inference_improvements}, for the sake of simplicity, for synthetic captions we choose the `l' and `l+i' captions, since we find that `l+e' and `l+u' have higher tendency to either reduce information (for `l+e') or else infer unnecessary detail (for `l+u').
Sampling short and medium length captions is less effective in this regime due to the information loss.
Introducing such ambiguity into the retrievals would be counterproductive.
To actually perform the retrieval, we compute the standard text-video similarity matrix, as well as a separate text-video similarity matrix for each type of 10k caption (`l' and `l+i').
We then add these together, giving $50\%$ weight to the standard text-video matrix, and equal weight to the remaining 2 matrices.

\subsection{Information Loss and Uniqueness Details}

\begin{figure*}[t]
\centering
    \begin{minipage}{0.475\linewidth}
    \centering
    \includegraphics[width=0.95\linewidth]{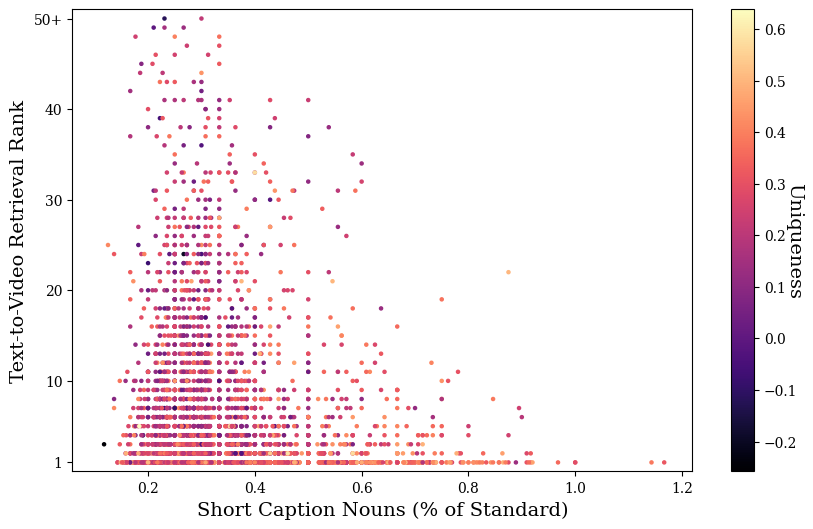}
    \caption{
        We measure the number of nouns and retrieval uniqueness for short caption retrieval, and find that the highest ranks correlate with captions that have lost their nouns and unique information.
    }
    \label{fig:rank_vs_entities_w_unique}
    \end{minipage}
    \hfill
    \begin{minipage}{0.475\linewidth}
    \centering
    \includegraphics[width=1.0\linewidth]{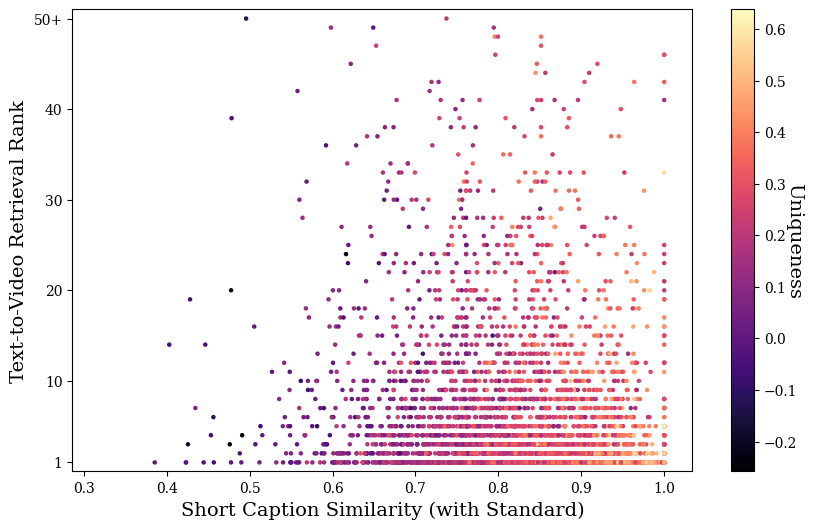}
    \caption{
        We measure the number of nouns and retrieval uniqueness for short caption retrieval, and find that the highest ranks correlate with captions that have lost their similarity with the source caption.
    }
    \label{fig:rank_vs_sim_w_unique}
    \end{minipage}
    \vspace{-1.0em}
\end{figure*}

We realize the length is not a perfect measure of information.
In fact, part of the motivation of this work is that captions can be quite short but very information-dense.
So, we compute information loss is 3 ways.
First, we use short length divided by standard length, as given in the main paper in Figure~\ref{fig:rank_vs_length_w_unique}.
Second, we use spaCy to count entities in the short and standard captions, dividing the number in the short by the number from the source standard caption in Figure~\ref{fig:rank_vs_entities_w_unique}.
Third, we get the word2vec embeddings for the entities in the short and standard captions, and compute the cosine similarities between all entities. We choose the best matches for the entries in the short caption, and sum the similarities, then divide by the number of entries in the short caption.
Hence we use similarity between bags of words as our proxy for how much the information in the short caption overlaps the information in the standard caption, with results in Figure~\ref{fig:rank_vs_sim_w_unique}.
These two alternatives confirm the findings from using length, so we opt to use length in the main paper since it is simpler.

To calculate uniqueness, we take the similarity score defined above (greedy matching of cosine similarities for word2vec embeddings of entities).
We additionally compute the similarity between the short caption and the standard captions for the top 5 retrieved videos, as retrieved using the short caption, not including the standard caption for the matching video.
That is, if the matching video is in the top 5 retrievals, we exclude it and additionally consider the standard caption for the video retrieved at rank 6.
We average the similarities between the short caption and these 5 standard captions, and subtract it from the similarity between short and source (matching) standard caption, for a uniqueness score. 
This ``uniqueness'' score provides the color in Figure~\ref{fig:rank_vs_length_w_unique}, Figure~\ref{fig:rank_vs_entities_w_unique}, and Figure~\ref{fig:rank_vs_sim_w_unique}.

\end{document}